\pdfoutput=1

\documentclass[11pt]{article}

\usepackage[]{naacl2021}

\usepackage{times}
\usepackage{latexsym}

\usepackage[T1]{fontenc}

\usepackage[utf8]{inputenc}
\usepackage{comment}
\usepackage{booktabs}
\usepackage{graphicx}
\usepackage{amsmath}
\usepackage[capitalize]{cleveref}
\usepackage{multirow,multicol}

\DeclareMathOperator*{\LeakyReLU}{\text{LeakyReLU}}
\DeclareMathOperator*{\softmax}{\text{softmax}}

\DeclareMathOperator*{\maxpool}{\text{maxpool}}
\DeclareMathOperator*{\GAT}{\text{GAT}}
\DeclareMathOperator*{\FFN}{\text{FFN}}

\title{Dialogue Relation Extraction with Document-Level \\ Heterogeneous Graph Attention Networks}

\author{Hui Chen, Pengfei Hong, Wei Han, Navonil Majumder, Soujanya Poria \\
  DeCLaRe Lab, Singapore University of Technology and Design, Singapore \\
  \texttt{hui\_chen@mymail.sutd.edu.sg,}\\
  \texttt{\{hongpengfei.emrys,henryhan88888\}@gmail.com,}\\
  \texttt{\{navonil\_majumder,sporia\}@sutd.edu.sg}\\}

\begin{document}
\maketitle
\begin{abstract}
Dialogue relation extraction (DRE) aims to detect the relation between pairs of entities mentioned in a multi-party dialogue. It plays an essential role in constructing knowledge graphs from conversational data increasingly abundant on the internet and facilitating intelligent dialogue system development. The prior methods of DRE do not meaningfully leverage speaker information---they just prepend the utterances with the respective speaker names. Thus, they fail to model the crucial inter-speaker relations that may provide additional context to relevant argument entities through pronouns and triggers. We present a graph attention network-based method for DRE where a graph that contains meaningfully connected speaker, entity, type, and utterance nodes is constructed. This graph is fed to a graph attention network for context propagation among relevant nodes, which effectively captures the dialogue context. We empirically show that this graph-based approach quite effectively captures the relations between different argument pairs in a dialogue as it outperforms the state-of-the-art approaches by a significant margin on the benchmark dataset DialogRE. Our code is released at: \url{https://github.com/declare-lab/dialog-HGAT}.
\end{abstract}

\section{Introduction}
The relation extraction (RE) task aims to identify relations between pairs of entities that exist in a document. It plays a pivotal role in understanding unstructured text and constructing knowledge bases~\cite{peng2017cross,quirk2017distant}.
Although the task of document-level relation extraction has been studied extensively in the past, the task of relation extraction from dialogues has yet to receive extensive study.

\begin{figure}[t]
\centering
\includegraphics[width=\columnwidth]{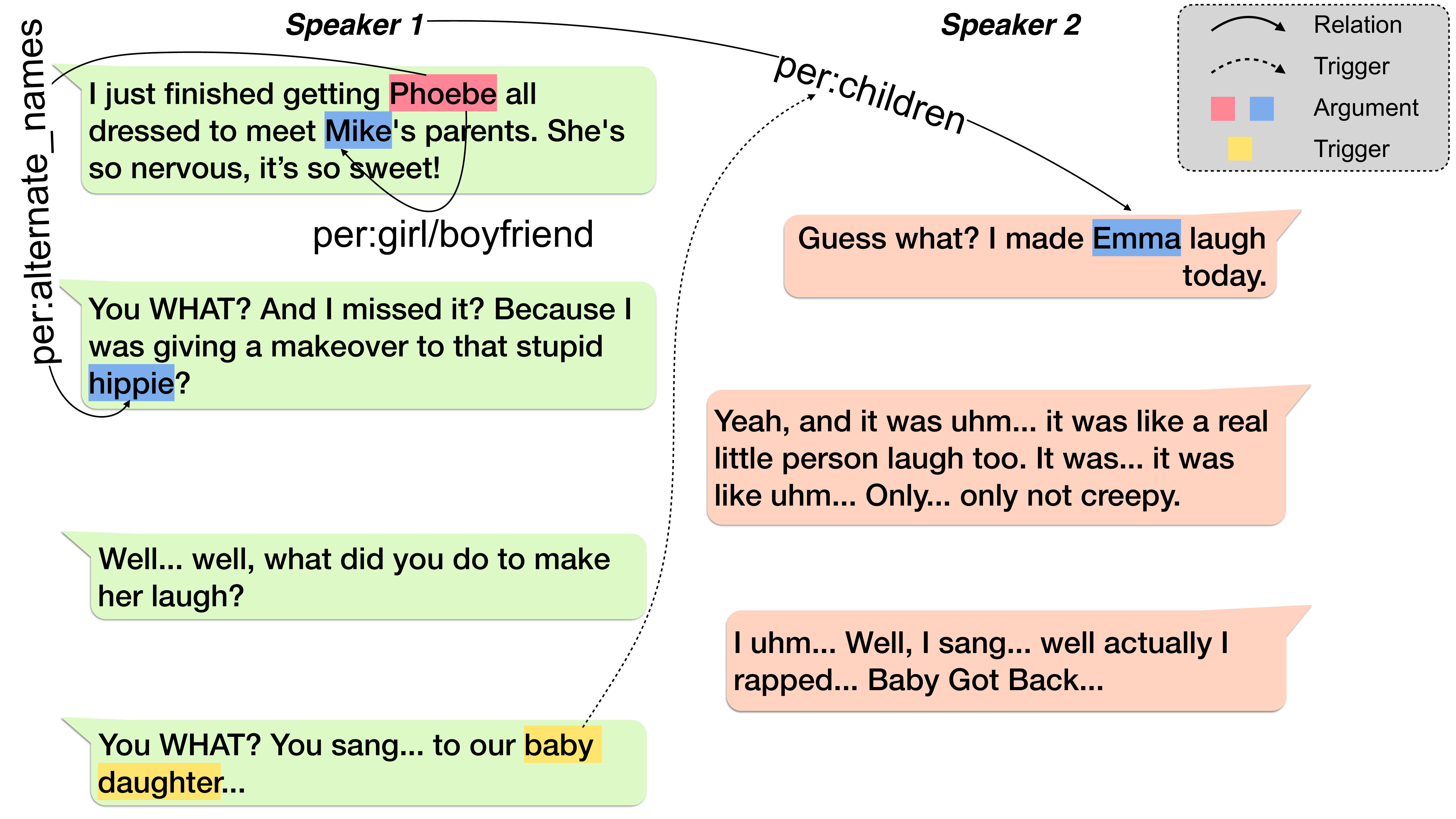}
\caption{An example adapted from DialogRE dataset. Words with red and blue background represent subject and object entities. Words with yellow background represent triggers that facilitate the relation inference. Solid and dash lines stand for intra- and inter-utterance relations.}
\label{fig1}
\end{figure}

Most previous works in this field focus on the professional and formal literature like biomedical documents~\cite{li2016biocreative,wu2019renet} and Wikipedia articles~\cite{elsahar2018t,yao2019docred,mesquita2019knowledgenet}.
These kinds of datasets are well-formatted and logically coherent with clear referential semantics.
Hence, for most NLP tasks, analyzing a few continuous sentences is enough to grasp pivotal information.
However, in dialogue relation extraction, conversational text is sampled from daily chat, which is more casual in nature.
Hence its logic is simpler but more entangled, and the referential ambiguity always occurs to an external reader.
Compared with formal literature, it has lower information density~\cite{wang2011pilot} and thus is more difficult for models to understand. Moreover, compared with other document-level RE datasets such as DocRED, dialogue text has much more cross-sentence relations~\cite{yu2020dialogue}.

\cref{fig1} presents an example of the target dialogue,
taken from DialogRE~\cite{yu2020dialogue} dataset. In order to infer the relation between \textit{Speaker1} and \textit{Emma}, we may need to find some triggers to recognize the characteristics of \textit{Emma}.
Triggers are shreds of evidence that can support the inference. As we can see, the following utterances are talking about \textit{Emma}, and the keyword \textit{baby daughter} mentioned by \textit{Speaker1} is a trigger, which provides evidence that \textit{Emma} is \textit{Speaker1}'s daughter.

Prior works show that triggers of arguments facilitate the document-level relation inference. Thus, the DocRED dataset~\cite{yao2019docred} provides several supporting evidence for each argument pair. Some efforts utilize the dependency paths of arguments to find possible triggers. For example, LSR model~\cite{nan2020reasoning} constructs meta dependency paths of each argument pair and aggregates all the word representations located in these paths to their model to enhance the model's reasoning ability.
\citet{sahu2019inter} uses syntactic parsing and coreference resolution to find intra- and inter-related words of each argument. \citet{christopoulou2019connecting} proposes an edge-oriented graph to synthesize argument-related information.
These models are graph-based and have proven powerful in encoding long-distance information. However, for dialogue relation extraction, interlocutors exist in every utterance of the dialogue, and they are often considered as an argument. Although these previous approaches have utilized entity features of arguments, most of them employ meta dependency paths to find the related words, which neglect necessary information related to speakers, since the speaker references have very little dependency features in each utterance. In this work, we formulate the dialogue relation extraction task as a classification problem, where we design a graph attention network to model semantic, syntactic, and speaker information. Compared with other graph-based models in the relation extraction task, our model is lightweight, without any costly matrix operation, and it can generalize to completely unseen graphs.

In this paper, we propose a simple yet effective attention-based heterogeneous graph neural network to tackle the dialogue relation extraction task in an inductive manner. We use multi-type features to create the graph and employ graph attention mechanism to propagate contextual information. Different from most of the previous works, our proposed model is customized for the relation extraction task in dialogue background, as we have specially modeled speaker information and designed a mechanism to propagate messages among different sentences for better inter-sentence representation learning.

The remainder of this paper is organized as follows: \cref{related-work} briefly discusses relevant works of heterogeneous graph neural networks; \cref{method} elaborates on our proposed framework; \cref{experiment} introduces the used dataset and baseline models; \cref{result} lays out the experiment results and analysis; \cref{conclusion} concludes the paper.

\section{Related Work}
\label{related-work}

Graph-based models have raised widespread attention from NLP researchers, as it is demonstrated as a powerful mathematical tool to represent complicated syntactic and semantic relations among structured language data.
Early work applies classic graph processing algorithms onto language graphs. \citet{pang2004sentimental} constructed a text graph and adopt the minimum-cut method to cluster the nodes for sentiment analysis. \citet{agirre2009personalizing} leveraged PageRank algorithm on personalized subgraphs of a wordnet to disambiguate polysemous words according to connected context words.

Recently, graph neural networks (GNN)~\cite{kipf2016semi} becomes popular in relation extraction tasks. For example, \citet{peng2017cross} tried to build a computation graph from syntactic parsing trees and employed graph LSTM to obtain better word embeddings for multi-ary relation extraction. \citet{zhang2018graph} designed a pruning algorithm for syntactic graphs and add a graph convolution layer on top of the sequential LSTM encoder in the learning process.
The combination with typical attention-based language models such as transformer~\cite{vaswani2017attention} is also studied.
\citet{cai2020graph} and~\citet{yao2020heterogeneous} used transformer-based graph convolutional networks to explicitly encode relations among distant syntactic nodes, to address the long-distance propagation issue.

Based on GNN, heterogeneous graph neural networks are proposed and have been applied in many NLP tasks, like text classification~\cite{linmei2019heterogeneous}, text summarization~\cite{wang2020heterogeneous}, user profiling~\cite{chen2019semi}, and event categorization~\cite{peng2019fine}. 
The prior work proves that heterogeneous graph neural network is a powerful tool in NLP. For the relation extraction task, \citet{christopoulou2019connecting} constructed an edge-oriented heterogeneous graph that contains sentence, mention, and entity information. However, syntactic information is neglected in their model. Differently, homogeneous nodes in our graph are all independent, and we take syntactic features to initialize sentence information as well as edges features.

\section{Method}
\label{method}
\subsection{Task Definition}
Given a dialogue containing $N$ utterances $\mathcal{D} = \{u_1, u_2, ..., u_N \}$ and a couple of argument pairs $\mathcal{A}=\{(x_1, y_1), (x_2,y_2),\dots\}$, where subject $x_i$ and object $y_i$ are entities mentioned in the dialogue, the goal is to identify the relation between argument pairs $(x_i, y_i)$. For document-level relation extraction task, there are many cross-sentence relations which are supported by various sentences.

\subsection{Model Overview}
\begin{figure*}[ht]
\centering
\includegraphics[width=\linewidth]{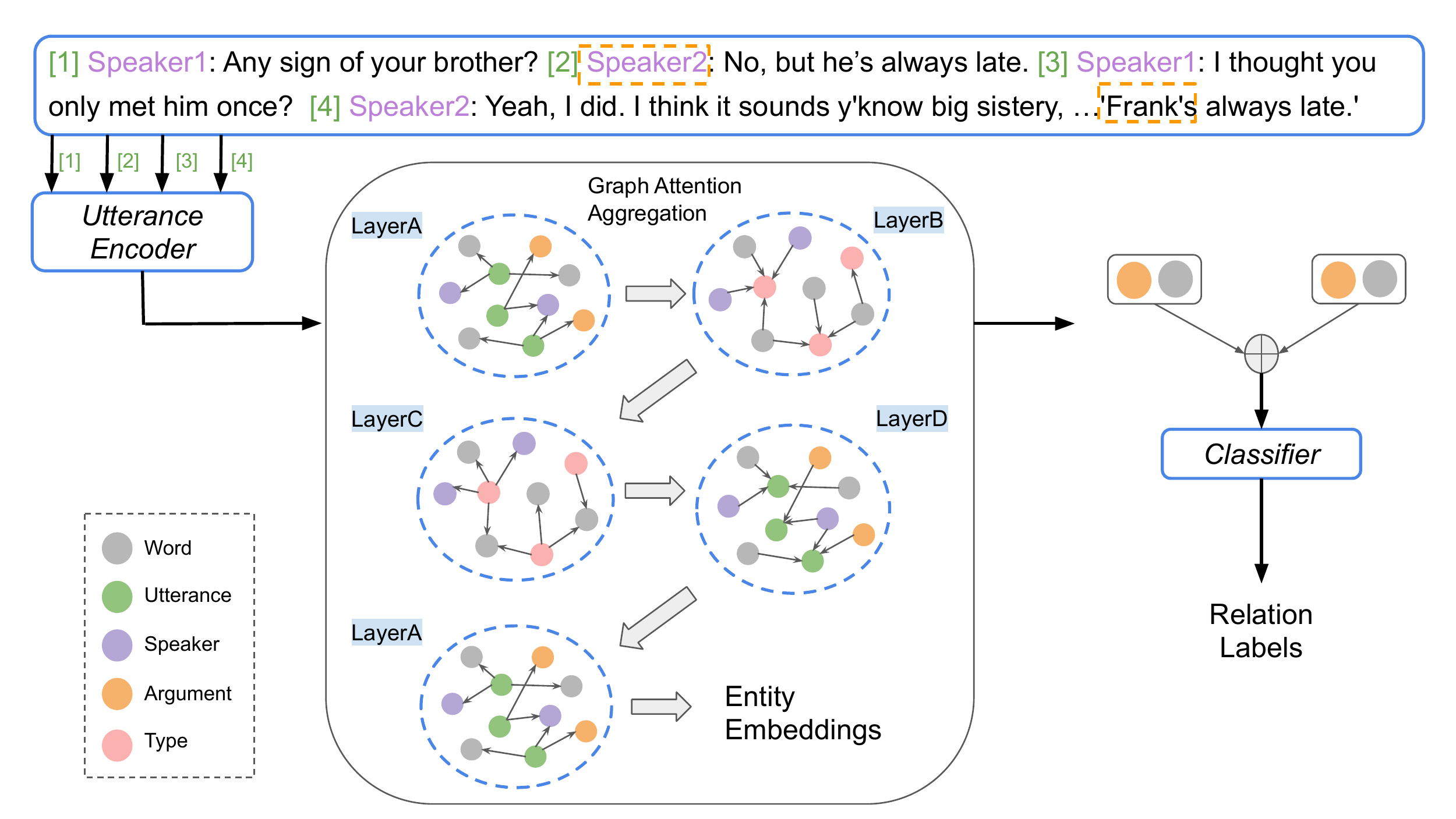}
\caption{An overview of the proposed model.}
\label{fig2}
\end{figure*}

In this work, we introduce an attention-based graph network to tackle the problem where each conversation is represented as a heterogeneous graph.
We first utilize an utterance encoder, which is composed of two  Bidirectional long short-term memory networks to encode conversational information.
These utterance encodings, along with word embeddings, speaker embeddings, argument embeddings, and type embedding, are logically connected to form a heterogeneous graph, which will be discussed in detail later in this section. Further, this graph is fed through five graph attention layers~\cite{velivckovic2017graph} that aggregate information from the neighboring nodes.
Lastly, we concatenate the learned argument embeddings and feed them to a classifier. An overview of the proposed model is shown in~\cref{fig2}.

\subsection{Data Preprocessing}
In the data preprocessing period, we use spaCy\footnote{\url{https://spacy.io}} to tokenize utterances, and at the same time, we obtain part-of-speech (POS) tags as well as named entity types of each token.

\subsection{Utterance Encoder}
Given a dialogue $\mathcal{D} = \{u_1, u_2, ..., u_N \}$, we use GloVe~\cite{pennington2014glove} to initialize the word embeddings and then feed them to a contextual Bidirectional Long Short-Term Memory network (BiLSTM) to obtain contextualized representations. The operation of BiLSTM can be defined as:
\begin{flalign}
    \label{lrnn1}
    \overleftarrow{h_{j}^{i}} &= LSTM_l(\overleftarrow{h_{j+1}^{i}}, e_{j}^{i}) \\
    \label{lrnn2}
    \overrightarrow{h_{j}^{i}} &= LSTM_r(\overrightarrow{h_{j-1}^{i}}, e_{j}^{i}) \\
    \label{lrnn3}
    h_{j}^{i} &= [\overleftarrow{h_{j}^{i}};  \overrightarrow{h_{j}^{i}}]
\end{flalign}
where $\overleftarrow{h_{j}^{i}}$ and $\overrightarrow{h_{j}^{i}}$ denote the hidden representations in the $j$-th layer of utterance $u_i$ from two directions, $h_{j}^{i}$ is the contextual representation which is the concatenation of $\overleftarrow{h_{j}^{i}}$ and $\overrightarrow{h_{j}^{i}}$, and $e_{j}^{i}$ stands for the embedding of the $j$-th token in utterance $u_{i}$. Unlike the previous approaches~\cite{christopoulou2019connecting,nan2020reasoning} that only adopt semantic contextual features in utterance encoding, we add syntactic features such as POS tags and named entity types to the contextual representations. The embedding of each token in the utterance can be described as:
\begin{flalign}
\label{utt-emb}
e = [e_{w}; e_{p}; e_{t}]
\end{flalign}
where we concatenate word embedding $e_{w}$ initialized by GloVe~\cite{pennington2014glove}, syntactic POS embedding $e_{p}$, and type embedding $e_{t}$ to form the token embedding $e$.

Moreover, we believe conversation-level contextual features play an important role in understanding a conversation. To encode non-local contextual information between each utterance, we apply max pool operation to the hidden states of each utterance-level BiLSTM (local LSTM), and then feed the sequence $c = \{c_1, c_2, ..., c_N \}$ to a conversation-level BiLSTM (global LSTM). The operation of global LSTM is the same as~\cref{lrnn1,lrnn2,lrnn3}.

\subsection{Graph Construction}

\subsubsection{Node Construction}
In our model, we design a heterogeneous graph network containing five types of nodes: utterance nodes, type
nodes, word nodes, speaker nodes, and argument nodes.
Each type of node is used to encode a type of information in the dialogue.
In the task, only word nodes, speaker nodes and argument nodes are probable to attend the final classification process. In other words, only these types of nodes are possible arguments.
For simplicity, we name them as \textit{basic nodes} in our illustration.

\paragraph{Utterance and Type Nodes}
Utterance nodes are initialized by the utterance embeddings which we obtain from the utterance encoder.
They are connected with the basic nodes which constitute the utterance.
Type nodes represent the entity types of words in an utterance, where a variety of named and numeric entities, such as PERSON or LOCATION, are included.
Since one mention may have different types in one conversation, type nodes can facilitate information integration. For example, `Frank' can be a string if it represents an alternative name, and at the same time, it can be a person if it refers to a speaker in the conversation. Type nodes are connected with the basic nodes having the type attribute in the conversation. Each type node is initialized with a specific type of embedding. We believe that type of information has a positive influence on the relation inference process.

\paragraph{Basic Nodes}
Word nodes represent the vocabulary of a conversation. Each word node is connected with the utterance, which contains the word and it is also connected with all the possible types that the word may have in the conversation. We initialize the states of word nodes with GloVe~\cite{pennington2014glove}.

Speaker nodes represent each unique speaker in the conversation. Each speaker node is connected with the utterances uttered by the speaker himself/herself. This type of node is initialized with some specific embeddings and can gather information from different speakers.

Argument nodes are two special nodes that are used to encode relative positional information of argument pairs. There are two argument nodes in each graph in total. One stands for the subject argument and the other represents the object argument. Similarly, argument nodes are also encoded by specific embeddings.

\subsubsection{Edge Construction}
The proposed graph is undirected but the propagation has directions. There are five types of edges: utterance-word, utterance-argument, utterance-speaker, type-word, and type-argument edges.  Each edge has its own type. These edges are randomly initialized except the utterance-word edge.

For the edge between utterance and word nodes, we adopt POS tags to initialize the edge features. This type of edge aggregates not only global semantic features of the conversation but also local syntactic features to the word nodes.

\subsubsection{Graph Attention Mechanism}
We use graph attention mechanism~\cite{velivckovic2017graph} to aggregate neighboring information to the target node. Suppose we have a node $i$ and some neighborhood nodes $j$,  the graph attention mechanism can be described as:

\begin{flalign}
\label{gat-global1}
\mathcal{F}(h_{i}, h_{j}) &= \LeakyReLU(\textbf{a}^{T}(\textbf{W}_{i}h_{i}; \textbf{W}_{j}h_{j}; \textbf{E}_{ij}) \\
\label{gat-global2} \alpha_{ij} &= \softmax(\mathcal{F}(h_{i}, h_{j})) \\
 \label{gat-g2}&= \frac{\exp(\mathcal{F}(h_{i}, h_{j}))}{\sum_{k}\exp(\mathcal{F}(h_{i}, h_{k}))} \\
\label{gat-global3} h_{i}' &= ||_{k=1}^{K}\sigma(\sum_{j}\alpha_{ij}^{k}\textbf{W}_{q}^{k}h_{j})
\end{flalign}
where ${h}_{i}$ and ${h}_{j}$ are representations of node $i$ and nodes $j$, $\textbf{W}_{i}$, $\textbf{W}_{j}$, $\textbf{W}_{q}$ and $\textbf{a}^T$ are trainable weight matrices, $\textbf{E}_{ij}$ is the edge weight matrix that is mapped to the multi-dimensional embedding space, $\alpha_{ij}$ is the attention weight between $i$ and $j$, $\sigma$ is an activation function, and $||$ is concatenation operation.

\subsubsection{Message Propagation}
As shown in~\cref{fig2}, there are five layers in our proposed graph module, where each layer represents an aggregation. There are four types of layers that we mark in the figure. LayerA and LayerD contain the message propagation between utterance nodes and basic nodes, and similarly, LayerB and LayerC are the message propagation between basic nodes and type nodes. We would call the whole message propagation path meta path. Different meta path strategies may lead to different performance.

Our meta path in this work can be described as follows:
First, we use utterance nodes to update word nodes, speaker nodes, and argument nodes; secondly, the updated word nodes and argument nodes pass messages to type nodes; then type nodes conversely update the word nodes and argument nodes; next we use word nodes, speaker nodes, and argument nodes to update utterance nodes; and lastly the updated utterance nodes update word nodes, speaker nodes and argument nodes. The path can be denoted as $V_{u}-V_{b}-V_{t}-V_{b}-V_{u}-V_{b}$, where $V_{u}$, $V_{b}$, and $V_{t}$ refer to utterance nodes, basic nodes, and type nodes.

Following~\citet{wang2020heterogeneous}, we add a residual connection~\cite{he2016deep} to avoid gradient vanishing during updating:

\begin{flalign}
\label{residual-global}
\hat{h_{i}} = \bar{h_{i}} + h_{i}'
\end{flalign}
where $\bar{h_{i}}$ is the output learned in the graph attention layer, and $h_{i}'$ is the original input of the graph attention layer.

In message passing, except for graph attention operation, there is also a two-layer feed-forward network which can be denoted as:
\begin{flalign}
\label{FFN}
h_{i}^{new} = \FFN(\hat{h_{i}})
\end{flalign}

Suppose we have the initial embeddings of utterance nodes, basic nodes and type nodes, denoted as embedding matrices $\mathbf{H}_u = \{\mathbf{H}_u, \mathbf{H}_b,  \mathbf{H}_t\}$, the message propagating process can be written as:

\begin{flalign}
\label{meta-path}
\mathbf{H}_b^1 &= \GAT(\mathbf{H}_b^0, \mathbf{H}_u^0) \\
\mathbf{H}_t^1 &= \GAT(\mathbf{H}_t^0, \mathbf{H}_b^1) \\
\mathbf{H}_b^2 &= \GAT(\mathbf{H}_b^1, \mathbf{H}_t^1) \\
\mathbf{H}_u^1 &= \GAT(\mathbf{H}_u^0, \mathbf{H}_b^2) \\
\mathbf{H}_b^3 &= \GAT(\mathbf{H}_b^2, \mathbf{H}_u^1)
\end{flalign}
where the GAT operation is the same as~\cref{gat-global1,gat-global2,gat-g2,gat-global3,residual-global,FFN}. The superscripts represent the $n^{th}$ update of the matrix and 0 marks the initial state.

\subsection{Relation Classifier}
After the message propagation in the heterogeneous graph, we obtain new representations of all entities. We select the argument nodes $\tau_x$ and $\tau_y$, as well as the corresponding word nodes $e_x$ and $e_y$ from basic nodes, and concatenate them. Finally, they are fed to a linear transformation and a sigmoid function to get the predictions:
\begin{flalign}
\label{classfier}
e_x' &= [\maxpool(\tau_{x}); \maxpool(e_{x})]  \\
e_y' &= [\maxpool(\tau_{y}); \maxpool(e_{y})]  \\
e' &= [e_{x}'; e_{y}']  \\
P(r|e_{x}, e_{y}) &= \sigma(\textbf{W}_{e}e'+b_{e})_{r}
\end{flalign}
where $P(r|e_{x}, e_{y})$ is the probability of relation type $r$ given argument pair $(e_{x}, e_{y})$, $\textbf{W}_{e}$ and $b_{e}$ are linear transformation weight and bias vector, $maxpool$ is max pooling operation, and $\sigma$ is sigmoid function.

\section{Experiments}
\label{experiment}
\subsection{Dataset Used}
We evaluate the proposed framework on the DialogRE dataset~\cite{yu2020dialogue}, which contains 1,788 dialogues and 10,168 relational triples. The data statistics are shown in~\cref{tab:dataset}. DialogRE is adapted from the complete transcripts of \textit{Friends}, a widely used corpus in dialogue research these years~\cite{chen-etal-2017-robust,zhou2018they,yang2019friendsqa,poria2019meld}, and there are 36 possible relation types, most of which focus on biographical attributes of person entities. Each dialogue contains several relational triples $(x,y,r)$, and the task is to predict the relation $r$ between each argument pair $(x,y)$. In the experiments, the dataset is partitioned into train, dev, and test set with a roughly 60/20/20 ratio. Following the evaluation metrics of DialogRE, we report macro $F1$ scores of the proposed model and all the baselines in both the standard and conversational settings. In the following sections, we use $F1_c$ to represent $F1$ scores in the conversational setting. 

\begin{table}[ht]
\centering
\begin{tabular}{l|c|c|c}
\toprule
 DialogRE & Train & Dev & Test \\
\midrule
\#Conversations & 1073 & 358 & 357 \\
\#Argument Pairs & 5963 & 1928 & 1858 \\
Average dialogue length & 229.5 & 224.1 & 214.2\\
Average \# of turns & 13.1 & 13.1 & 12.4 \\
Average \# of speakers & 3.3 & 3.2 & 3.3 \\
\bottomrule
\end{tabular}
\caption{DialogRE dataset statistics.}
\label{tab:dataset}
\end{table}

\subsection{Baseline models}
\subsubsection{Sequence-based Models}
We select convolutional neural networks (CNN)~\cite{zeng2014relation}, LSTM, and BiLSTM~\cite{cai2016bidirectional} as the sequence-based baselines. These models take word embeddings, mention embeddings, and type embeddings as features. Concretely, they use GloVe and spaCy to get word embeddings and label named-entity types, and then take an average of all the embeddings of mention names for each entity to get mention embeddings.

\subsubsection{Graph-based Models}
As our proposed model is graph-based, we also select two graph-based models AGGCN~\cite{guo2019attention} and LSR~\cite{nan2020reasoning} as the baselines. AGGCN directly feeds the full dependency tree of each sentence to a graph convolutional network, which takes self-attention weights as soft edges. It achieves state-of-the-art results in various relation extraction tasks. LSR adopts an adaptation of Kirchhoff’s Matrix-Tree Theorem~\cite{tutte1984graph,koo2007structured} to induce the latent dependency structure of each document and then feeds the latent structure to a densely connected graph convolutional network to inference the relations. These graph-based models both utilize dependency information to construct the inference graph.

\section{Result and Analysis}
\label{result}
\subsection{Comparison with Baselines}
We present our main results on DialogRE dataset in~\cref{tab:results}. As shown in~\cref{tab:results}, our model surpasses the state-of-the-art method by 9.6\%/7.5\% $F1$ scores, and 8.4\%/5.7\% $F1_c$ scores in both validation and test sets, which demonstrates the effectiveness of the information propagation along task-specific functional meta-paths in the heterogeneous graph.
As a result, inter-sentence communication usually passes through a long distance, which causes information loss or degradation.
However, this kind of information transmission is critically important for dialog-style text, because logical connections are not locally compact within adjacent sentences,
instead, they are spread over the whole conversations.
Our proposed model constructs a heterogeneous graph with shorter distances between logically closed but syntactically faraway word pairs.
Hence the long-distance issue is mitigated.
\par
We also compare the model sizes as an efficiency indicator.
Although creating numerous nodes and edges inevitably brings overhead, the total number of parameters is still moderate.

\begin{table*}[ht]
\centering
\begin{tabular}{l|c|cccc}
\toprule
\multirow{2}{*}{Model} & &
\multicolumn{2}{c}{Dev (\%)} &
\multicolumn{2}{c}{Test (\%)} \\
  & \#params & $F1$ & $F1_c$ & $F1$ & $F1_c$ \\
\midrule
Majority~\cite{yu2020dialogue} & - & 38.9 & 38.7 & 35.8 & 35.8 \\
CNN~\cite{yu2020dialogue} & - & 46.1 & 43.7 & 48.0 & 45.0 \\
LSTM~\cite{yu2020dialogue} & - & 46.7 & 44.2 & 47.4 & 44.9 \\
BiLSTM~\cite{yu2020dialogue} & 4.1M & 48.1 & 44.3 & 48.6 & 45.0 \\
\midrule
AGGCN~\cite{guo2019attention} & 3.7M & 46.6 & 40.5 & 46.2 & 39.5 \\
LSR~\cite{nan2020reasoning} & 20.5M & 44.5 & - & 44.4 & - \\
\midrule
This work & 4.0M & \textbf{57.7} & \textbf{52.7} & \textbf{56.1} & \textbf{50.7} \\
\bottomrule
\end{tabular}
\caption{Main results on DialogRE dataset. Values in the \#params column refer to parameter sizes of the models. $F1$ and $F1_c$ are macro $F1$ scores under standard setting and conversational setting, respectively. Word embeddings of the models are captured by GloVe~\cite{pennington2014glove}.}
\label{tab:results}
\end{table*}

\subsection{Ablation Study}
To understand the impact of our model's components, we perform ablation studies using our proposed model on the DialogRE dataset. The ablation results are shown in~\cref{tab:ablation}. First, we remove local LSTM and global LSTM.
The dropping accuracy
proves that the contextual encoder plays an important role in semantic feature extraction. Second, we remove the specific argument nodes and have observed that $F1$ and ${F1_c}$ scores decrease to 55.0\% and 50.2\% on test set. This proves that our design on argument nodes effectively synthesizes argument features to the model.
Further, we test the performance of the syntactic features we inject by removing POS embedding, NER embedding, and POS edge features.
The scores record a decrease under all these experiment settings. Notably, removing POS embedding leads to even about 2\% drops in all the evaluation metrics.

\begin{table}[ht]
\centering
\small
\begin{tabular}{l|cccc}
\toprule
\multirow{2}{*}{Model} &
\multicolumn{2}{c}{Dev (\%)} &
\multicolumn{2}{c}{Test (\%)} \\
  & $F1$ & $F1_c$ & $F1$ & $F1_c$ \\
\midrule
Full model & 57.7 & 52.7 & 56.1 & 50.7 \\
w/o Local BiLSTM & 54.9 & 50.0 & 55.3 & 50.3 \\
w/o Global BiLSTM & 54.7 & 50.2  & 53.5 & 48.7 \\
w/o Argument nodes & 56.0 & 51.3 & 55.0 & 50.2 \\
w/o POS embedding & 54.6 & 50.9 & 53.0 & 48.5   \\
w/o NER embedding & 56.8 & 51.5 & 54.2 & 49.2 \\
w/o POS edge weights & 56.9 & 52.4 & 54.7 & 50.4 \\

\bottomrule
\end{tabular}
\caption{Ablation results on DialogRE dataset. }
\label{tab:ablation}
\end{table}

\subsection{Effect of the Meta Path}
We test the performance of our message propagation strategy via changing meta-path strategies. In our proposed model, there are five layers in the heterogeneous graph.
Those basic nodes, corresponding to different types of words, speakers, and arguments, are updated totally three times, i.e., they are first updated by utterance nodes, second updated by type nodes, and ultimately updated by utterance again.
To investigate the meta path's effect, we compare our proposed five-layer graph module with different strategies where the numbers of layers are one, seven, and nine in~\cref{tab:effect}.
In Strategy1, we only set up one LayerA, where the basic nodes are updated by the initialized utterance nodes once.
We observe that all the macro $F1$ scores drop dramatically, showing the one-layer structure is not deep enough to grasp complex dependencies. To make node features more informative, we would add more layers.
At this time, we may be curious about how many layers the module should have to induct an optimal structure in this task.
In Strategy2 and Strategy3, we design a seven-layer module and a nine-layer module, respectively. For Strategy2, the order of layers is A-B-C-D-A-D-A, where A,B,C, and D are layer labels introduced in~\cref{fig2}. Compared with our proposed module, scores on validation set decrease about 1\% and scores on test set decrease 1.7\% and 0.6 \% with the standard-setting and the conversational setting, respectively.
However, the module with nine layers in Strategy3 shows a larger gap between itself and the best performance, where the order of layers is A-B-C-D-A-B-C-D-A. We think this is probably because the structure is so complicated, which causes an over-smooth problem and prevents itself from learning meaningful hidden representations.

\begin{table}[ht]
\centering
\small
\begin{tabular}{l|cccc}
\toprule
\multirow{2}{*}{Strategy} &
\multicolumn{2}{c}{Dev (\%)} &
\multicolumn{2}{c}{Test (\%)} \\
  & $F1$ & $F1_c$ & $F1$ & $F1_c$ \\
\midrule
 This work (L=5) & 57.7 & 52.7 & 56.1 & 50.7 \\
 Strategy1 (L=1) & 48.3 & 46.5 & 48.4 & 45.9 \\
 Strategy2 (L=7) & 56.8 & 51.6 & 54.4 & 50.1 \\
 Strategy3 (L=9) & 53.8 & 49.1 & 52.2 & 47.2 \\
\bottomrule
\end{tabular}
\caption{Comparison with different meta-path strategies on DialogRE dataset. `L' means the number of layers in the graph module. }
\label{tab:effect}
\end{table}

\subsection{Case Studies}
\begin{figure}[t]
\centering
\includegraphics[width=0.9\columnwidth]{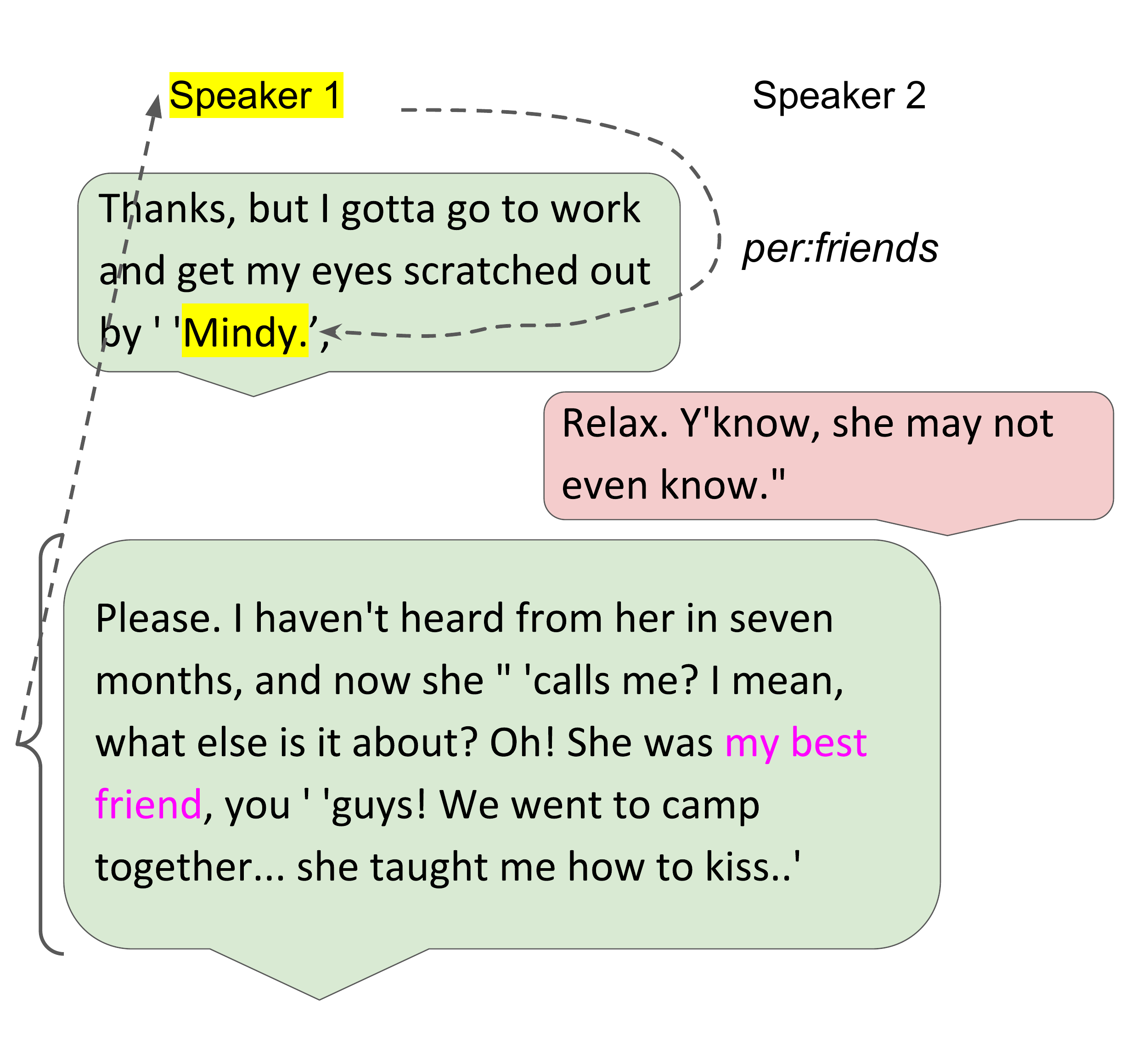} 
\caption{An example to show the effective message propagation between argument pairs}
\label{case_study}
\end{figure}

In the dataset, 95\% of argument pairs span in at least two consecutive sentences instead of being restricted to the same sentence.
Therefore, it is crucial that the model can tackle long-distance learning issues.
Compared with the LSTM model,
direct connections among different types of nodes in HGNN reduce the length of information propagation paths between pairs of argument nodes.
Considering the following example in \cref{case_study}, subject a - `Mindy' and object b - `Speaker 1' share the relationship `per:friends', which is indicated by the trigger `my best friend' in the first utterance.
The entity information is relayed from `Mindy' to `Speaker 1' in the update process: `speaker 1' node aggregates utterance level information from its neighbor nodes containing a. the relation trigger `best friend'. b. in BiLSTM model, the key information has to travel a long journey from the subject entity word to the object one as there are too many words between them in the context.

\subsection{Error Analysis}
Type information involves in the information propagation process and thus affects the contents of output embeddings.
The model is prone to make incorrectly and biased predictions.
If it fails to receive enough certainty from other information sources and then can only rely on the entity types of the two arguments.
For example, given an argument pair of two human names, both are named entity type `PERSON'.
Sometimes the model inclines to deem the relationship between the two arguments to be `per:alternate\_name' instead of the correct answer `per:alumni' or `per:roommate'.
This is because among all of these classes, `PERSON-PERSON' is a preferable type pair.
However, the class `per:alternate\_name' (22.01\%) presents more frequently than `per:alumni' (1.83\%) and `per:roommate' (1.29\%) in the dataset. 
When information aggregated from all sources other than the argument pair is not evident for judgment, entity bias misguides the model to the wrong classification results.

\section{Conclusion}
\label{conclusion}
In this work, we present an attention-based heterogeneous graph network to deal with the dialogue relation extraction task in an inductive manner. This heterogeneous graph attention network has modeled multi-type features of the conversation, such as utterance, word, speaker, argument, and entity type information. On the benchmark DialogRE dataset, our proposed framework outperforms the strongest baselines and the state-of-the-art approaches by a significant margin, which proves the proposed framework can effectively capture relations between different entities in the conversation. Future work will focus on making use of latent relations between entities that exist in dialogue history to develop intelligent conversational agents.

\bibliography{main}

\begin{thebibliography}{35}
\expandafter\ifx\csname natexlab\endcsname\relax\def\natexlab#1{#1}\fi

\bibitem[{Agirre and Soroa(2009)}]{agirre2009personalizing}
Eneko Agirre and Aitor Soroa. 2009.
\newblock Personalizing pagerank for word sense disambiguation.
\newblock In \emph{Proceedings of the 12th Conference of the European Chapter
  of the ACL (EACL 2009)}, pages 33--41.

\bibitem[{Cai and Lam(2020)}]{cai2020graph}
Deng Cai and Wai Lam. 2020.
\newblock Graph transformer for graph-to-sequence learning.
\newblock In \emph{AAAI}, pages 7464--7471.

\bibitem[{Cai et~al.(2016)Cai, Zhang, and Wang}]{cai2016bidirectional}
Rui Cai, Xiaodong Zhang, and Houfeng Wang. 2016.
\newblock Bidirectional recurrent convolutional neural network for relation
  classification.
\newblock In \emph{Proceedings of the 54th Annual Meeting of the Association
  for Computational Linguistics (Volume 1: Long Papers)}, pages 756--765.

\bibitem[{Chen et~al.(2017)Chen, Zhou, and Choi}]{chen-etal-2017-robust}
Henry~Y. Chen, Ethan Zhou, and Jinho~D. Choi. 2017.
\newblock Robust coreference resolution and entity linking on dialogues:
  Character identification on {TV} show transcripts.
\newblock In \emph{Proceedings of the 21st Conference on Computational Natural
  Language Learning ({C}o{NLL} 2017)}, pages 216--225, Vancouver, Canada.
  Association for Computational Linguistics.

\bibitem[{Chen et~al.(2019)Chen, Gu, Ren, He, Xie, Guo, Yin, and
  Zhang}]{chen2019semi}
Weijian Chen, Yulong Gu, Zhaochun Ren, Xiangnan He, Hongtao Xie, Tong Guo,
  Dawei Yin, and Yongdong Zhang. 2019.
\newblock Semi-supervised user profiling with heterogeneous graph attention
  networks.
\newblock In \emph{IJCAI}, volume~19, pages 2116--2122.

\bibitem[{Christopoulou et~al.(2019)Christopoulou, Miwa, and
  Ananiadou}]{christopoulou2019connecting}
Fenia Christopoulou, Makoto Miwa, and Sophia Ananiadou. 2019.
\newblock Connecting the dots: Document-level neural relation extraction with
  edge-oriented graphs.
\newblock In \emph{Proceedings of the 2019 Conference on Empirical Methods in
  Natural Language Processing and the 9th International Joint Conference on
  Natural Language Processing (EMNLP-IJCNLP)}, pages 4927--4938.

\bibitem[{Elsahar et~al.(2018)Elsahar, Vougiouklis, Remaci, Gravier, Hare,
  Laforest, and Simperl}]{elsahar2018t}
Hady Elsahar, Pavlos Vougiouklis, Arslen Remaci, Christophe Gravier, Jonathon
  Hare, Frederique Laforest, and Elena Simperl. 2018.
\newblock T-rex: A large scale alignment of natural language with knowledge
  base triples.
\newblock In \emph{Proceedings of the Eleventh International Conference on
  Language Resources and Evaluation (LREC 2018)}.

\bibitem[{Guo et~al.(2019)Guo, Zhang, and Lu}]{guo2019attention}
Zhijiang Guo, Yan Zhang, and Wei Lu. 2019.
\newblock Attention guided graph convolutional networks for relation
  extraction.
\newblock In \emph{Proceedings of the 57th Annual Meeting of the Association
  for Computational Linguistics}, pages 241--251.

\bibitem[{He et~al.(2016)He, Zhang, Ren, and Sun}]{he2016deep}
Kaiming He, Xiangyu Zhang, Shaoqing Ren, and Jian Sun. 2016.
\newblock Deep residual learning for image recognition.
\newblock In \emph{Proceedings of the IEEE conference on computer vision and
  pattern recognition}, pages 770--778.

\bibitem[{Kipf and Welling(2017)}]{kipf2016semi}
Thomas~N Kipf and Max Welling. 2017.
\newblock Semi-supervised classification with graph convolutional networks.
\newblock In \emph{5th International Conference on Learning Representations}.

\bibitem[{Koo et~al.(2007)Koo, Globerson, Carreras, and
  Collins}]{koo2007structured}
Terry Koo, Amir Globerson, Xavier Carreras, and Michael Collins. 2007.
\newblock Structured prediction models via the matrix-tree theorem.
\newblock In \emph{Proceedings of the 2007 Joint Conference on Empirical
  Methods in Natural Language Processing and Computational Natural Language
  Learning (EMNLP-CoNLL)}, pages 141--150.

\bibitem[{Li et~al.(2016)Li, Sun, Johnson, Sciaky, Wei, Leaman, Davis,
  Mattingly, Wiegers, and Lu}]{li2016biocreative}
Jiao Li, Yueping Sun, Robin~J Johnson, Daniela Sciaky, Chih-Hsuan Wei, Robert
  Leaman, Allan~Peter Davis, Carolyn~J Mattingly, Thomas~C Wiegers, and Zhiyong
  Lu. 2016.
\newblock Biocreative v cdr task corpus: a resource for chemical disease
  relation extraction.
\newblock \emph{Database}, 2016.

\bibitem[{Linmei et~al.(2019)Linmei, Yang, Shi, Ji, and
  Li}]{linmei2019heterogeneous}
Hu~Linmei, Tianchi Yang, Chuan Shi, Houye Ji, and Xiaoli Li. 2019.
\newblock Heterogeneous graph attention networks for semi-supervised short text
  classification.
\newblock In \emph{Proceedings of the 2019 Conference on Empirical Methods in
  Natural Language Processing and the 9th International Joint Conference on
  Natural Language Processing (EMNLP-IJCNLP)}, pages 4823--4832.

\bibitem[{Mesquita et~al.(2019)Mesquita, Cannaviccio, Schmidek, Mirza, and
  Barbosa}]{mesquita2019knowledgenet}
Filipe Mesquita, Matteo Cannaviccio, Jordan Schmidek, Paramita Mirza, and
  Denilson Barbosa. 2019.
\newblock Knowledgenet: A benchmark dataset for knowledge base population.
\newblock In \emph{Proceedings of the 2019 Conference on Empirical Methods in
  Natural Language Processing and the 9th International Joint Conference on
  Natural Language Processing (EMNLP-IJCNLP)}, pages 749--758.

\bibitem[{Nan et~al.(2020)Nan, Guo, Sekulic, and Lu}]{nan2020reasoning}
Guoshun Nan, Zhijiang Guo, Ivan Sekulic, and Wei Lu. 2020.
\newblock Reasoning with latent structure refinement for document-level
  relation extraction.
\newblock In \emph{Proceedings of the 58th Annual Meeting of the Association
  for Computational Linguistics}, pages 1546--1557.

\bibitem[{Pang and Lee(2004)}]{pang2004sentimental}
Bo~Pang and Lillian Lee. 2004.
\newblock A sentimental education: Sentiment analysis using subjectivity
  summarization based on minimum cuts.
\newblock In \emph{Proceedings of the 42nd Annual Meeting of the Association
  for Computational Linguistics (ACL-04)}, pages 271--278.

\bibitem[{Peng et~al.(2019)Peng, Li, Gong, Song, Ning, Lai, and
  Yu}]{peng2019fine}
Hao Peng, Jianxin Li, Qiran Gong, Yangqiu Song, Yuanxing Ning, Kunfeng Lai, and
  Philip~S Yu. 2019.
\newblock Fine-grained event categorization with heterogeneous graph
  convolutional networks.
\newblock In \emph{Proceedings of the 28th International Joint Conference on
  Artificial Intelligence}, pages 3238--3245. AAAI Press.

\bibitem[{Peng et~al.(2017)Peng, Poon, Quirk, Toutanova, and
  Yih}]{peng2017cross}
Nanyun Peng, Hoifung Poon, Chris Quirk, Kristina Toutanova, and Wen-tau Yih.
  2017.
\newblock Cross-sentence n-ary relation extraction with graph lstms.
\newblock \emph{Transactions of the Association for Computational Linguistics},
  5:101--115.

\bibitem[{Pennington et~al.(2014)Pennington, Socher, and
  Manning}]{pennington2014glove}
Jeffrey Pennington, Richard Socher, and Christopher~D Manning. 2014.
\newblock Glove: Global vectors for word representation.
\newblock In \emph{Proceedings of the 2014 conference on empirical methods in
  natural language processing (EMNLP)}, pages 1532--1543.

\bibitem[{Poria et~al.(2019)Poria, Hazarika, Majumder, Naik, Cambria, and
  Mihalcea}]{poria2019meld}
Soujanya Poria, Devamanyu Hazarika, Navonil Majumder, Gautam Naik, Erik
  Cambria, and Rada Mihalcea. 2019.
\newblock Meld: A multimodal multi-party dataset for emotion recognition in
  conversations.
\newblock In \emph{Proceedings of the 57th Annual Meeting of the Association
  for Computational Linguistics}, pages 527--536.

\bibitem[{Quirk and Poon(2017)}]{quirk2017distant}
Chris Quirk and Hoifung Poon. 2017.
\newblock Distant supervision for relation extraction beyond the sentence
  boundary.
\newblock In \emph{Proceedings of the 15th Conference of the European Chapter
  of the Association for Computational Linguistics: Volume 1, Long Papers},
  pages 1171--1182.

\bibitem[{Sahu et~al.(2019)Sahu, Christopoulou, Miwa, and
  Ananiadou}]{sahu2019inter}
Sunil~Kumar Sahu, Fenia Christopoulou, Makoto Miwa, and Sophia Ananiadou. 2019.
\newblock Inter-sentence relation extraction with document-level graph
  convolutional neural network.
\newblock In \emph{Proceedings of the 57th Annual Meeting of the Association
  for Computational Linguistics}, pages 4309--4316.

\bibitem[{Tutte(1984)}]{tutte1984graph}
William~Thomas Tutte. 1984.
\newblock Graph theory.
\newblock In \emph{Claren- don Press}.

\bibitem[{Vaswani et~al.(2017)Vaswani, Shazeer, Parmar, Uszkoreit, Jones,
  Gomez, Kaiser, and Polosukhin}]{vaswani2017attention}
Ashish Vaswani, Noam Shazeer, Niki Parmar, Jakob Uszkoreit, Llion Jones,
  Aidan~N Gomez, {\L}ukasz Kaiser, and Illia Polosukhin. 2017.
\newblock Attention is all you need.
\newblock In \emph{Advances in neural information processing systems}, pages
  5998--6008.

\bibitem[{Veli{\v{c}}kovi{\'c} et~al.(2018)Veli{\v{c}}kovi{\'c}, Cucurull,
  Casanova, Romero, Lio, and Bengio}]{velivckovic2017graph}
Petar Veli{\v{c}}kovi{\'c}, Guillem Cucurull, Arantxa Casanova, Adriana Romero,
  Pietro Lio, and Yoshua Bengio. 2018.
\newblock Graph attention networks.
\newblock In \emph{6th International Conference on Learning Representations}.

\bibitem[{Wang et~al.(2020)Wang, Liu, Zheng, Qiu, and
  Huang}]{wang2020heterogeneous}
Danqing Wang, Pengfei Liu, Yining Zheng, Xipeng Qiu, and Xuanjing Huang. 2020.
\newblock Heterogeneous graph neural networks for extractive document
  summarization.
\newblock In \emph{Proceedings of the 58th Annual Meeting of the Association
  for Computational Linguistics}, pages 6209--6219.

\bibitem[{Wang and Liu(2011)}]{wang2011pilot}
Dong Wang and Yang Liu. 2011.
\newblock A pilot study of opinion summarization in conversations.
\newblock In \emph{Proceedings of the 49th annual meeting of the Association
  for Computational Linguistics: Human language technologies}, pages 331--339.

\bibitem[{Wu et~al.(2019)Wu, Luo, Leung, Ting, and Lam}]{wu2019renet}
Ye~Wu, Ruibang Luo, Henry~CM Leung, Hing-Fung Ting, and Tak-Wah Lam. 2019.
\newblock Renet: A deep learning approach for extracting gene-disease
  associations from literature.
\newblock In \emph{International Conference on Research in Computational
  Molecular Biology}, pages 272--284. Springer.

\bibitem[{Yang and Choi(2019)}]{yang2019friendsqa}
Zhengzhe Yang and Jinho~D Choi. 2019.
\newblock Friendsqa: Open-domain question answering on tv show transcripts.
\newblock In \emph{Proceedings of the 20th Annual SIGdial Meeting on Discourse
  and Dialogue}, pages 188--197.

\bibitem[{Yao et~al.(2020)Yao, Wang, and Wan}]{yao2020heterogeneous}
Shaowei Yao, Tianming Wang, and Xiaojun Wan. 2020.
\newblock Heterogeneous graph transformer for graph-to-sequence learning.
\newblock In \emph{Proceedings of the 58th Annual Meeting of the Association
  for Computational Linguistics}, pages 7145--7154.

\bibitem[{Yao et~al.(2019)Yao, Ye, Li, Han, Lin, Liu, Liu, Huang, Zhou, and
  Sun}]{yao2019docred}
Yuan Yao, Deming Ye, Peng Li, Xu~Han, Yankai Lin, Zhenghao Liu, Zhiyuan Liu,
  Lixin Huang, Jie Zhou, and Maosong Sun. 2019.
\newblock Docred: A large-scale document-level relation extraction dataset.
\newblock In \emph{Proceedings of the 57th Annual Meeting of the Association
  for Computational Linguistics}, pages 764--777.

\bibitem[{Yu et~al.(2020)Yu, Sun, Cardie, and Yu}]{yu2020dialogue}
Dian Yu, Kai Sun, Claire Cardie, and Dong Yu. 2020.
\newblock Dialogue-based relation extraction.
\newblock In \emph{Proceedings of the 58th Annual Meeting of the Association
  for Computational Linguistics}, pages 4927--4940.

\bibitem[{Zeng et~al.(2014)Zeng, Liu, Lai, Zhou, and Zhao}]{zeng2014relation}
Daojian Zeng, Kang Liu, Siwei Lai, Guangyou Zhou, and Jun Zhao. 2014.
\newblock Relation classification via convolutional deep neural network.
\newblock In \emph{Proceedings of COLING 2014, the 25th International
  Conference on Computational Linguistics: Technical Papers}, pages 2335--2344.

\bibitem[{Zhang et~al.(2018)Zhang, Qi, and Manning}]{zhang2018graph}
Yuhao Zhang, Peng Qi, and Christopher~D Manning. 2018.
\newblock Graph convolution over pruned dependency trees improves relation
  extraction.
\newblock In \emph{Proceedings of the 2018 Conference on Empirical Methods in
  Natural Language Processing}, pages 2205--2215.

\bibitem[{Zhou and Choi(2018)}]{zhou2018they}
Ethan Zhou and Jinho~D Choi. 2018.
\newblock They exist! introducing plural mentions to coreference resolution and
  entity linking.
\newblock In \emph{Proceedings of the 27th International Conference on
  Computational Linguistics}, pages 24--34.

\end{thebibliography}
\bibliographystyle{acl_natbib}

\clearpage

\appendix

\section{Settings and Hyperparameters}
\label{sec:appendix1}
In our experiments, we tune the parameters of batch size, learning rate, and BiLSTM hidden size by testing the performance on the validation set. \cref{tab:param_settings} lists the major parameters used in our experiments.

\begin{table}[ht]
\centering
\begin{tabular}{l|c}
\toprule
 Parameter & Value \\
\midrule
Word embedding dimension & 300 \\
NER embedding dimension & 30 \\
POS embedding dimension & 30 \\
Local BiLSTM hidden Size & 200 \\
Local BiLSTM layers & 2 \\
Global BiLSTM hidden Size & 128 \\
Global BiLSTM layers & 2 \\
\# Multihead attention & 10 \\
Learning rate & 0.0005 \\
Batch size & 16 \\
Edge embedding dimension & 50 \\
\bottomrule
\end{tabular}
\caption{Parameter settings.}
\label{tab:param_settings}
\end{table}

\section{Statistics of Relation Labels}
\label{sec:appendix2}
\cref{tab:relation-labels} shows statistics of relation labels in DialogRE dataset. In the train set and test set, there are 35 types of relations, while in the dev set, there are 37 types. `gpe:birth\_in\_place' and `per:place\_of\_birth' only exist in the dev set.

\begin{table*}[ht]
\centering

\begin{tabular}{l|ccc|ccc}
\toprule
 \multirow{2}{*}{Relation Type} & \multicolumn{3}{c}{Quantity} & \multicolumn{3}{c}{Percentage (\%)} \\
  & train & dev & test & train & dev & test \\
\midrule
per:alternate\_names & 1319 & 410 & 409 & 22.12 & 21.26 & 22.01  \\
unanswerable & 1308 & 404 & 388 & 21.94 & 20.95 & 20.88 \\
per:girl/boyfriend & 502 & 170 & 136 & 8.42 & 8.82 & 7.32 \\
per:positive\_impression & 476 & 149 & 138 & 7.98 & 7.73 & 7.43 \\
per:friends & 444 & 156 & 122 & 7.45 & 8.09 & 6.57 \\
per:title & 250 & 86 & 78 & 4.19 & 4.46 & 4.20 \\
per:spouse & 204 & 72 & 54 & 3.42 & 3.73 & 2.91 \\
per:siblings & 196 & 64 & 58 & 3.29 & 3.32 & 3.12 \\
per:children & 171 & 55 & 48 & 2.87 & 2.85 & 2.58 \\
per:parents & 171 & 55 & 48 & 2.87 & 2.85 & 2.58 \\
per:negative\_impression & 156 & 46 & 56 & 2.62 & 2.39 & 3.01 \\
per:roommate & 140 & 44 & 24 & 2.35 & 2.28 & 1.29 \\
per:alumni & 110 & 38 & 34 & 1.84 & 1.97 & 1.83 \\
per:other\_family & 66 & 29 & 30 & 1.11 & 1.50 & 1.61 \\
per:works & 58 & 12 & 19 & 0.97 & 0.62 & 1.02 \\
per:age & 53 & 15 & 10 & 0.89 & 0.78 & 0.54 \\
per:client & 52 & 18 & 18 & 0.87 & 0.93 & 0.97 \\
per:place\_of\_residence & 49 & 12 & 23 & 0.82 & 0.62 & 1.24 \\
gpe:residents\_of\_place & 49 & 12 & 23 & 0.82 & 0.62 & 1.24 \\
per:boss & 49 & 13 & 12 & 0.82 & 0.67 & 0.65 \\
per:subordinate & 49 & 13 & 12 & 0.82 & 0.67 & 0.65 \\
per:visited\_place & 48 & 20 & 25 & 0.80 & 1.04 & 1.35 \\
gpe:visitors\_of\_place & 48 & 20 & 25 & 0.80 & 1.04 & 1.35 \\
per:employee\_or\_member\_of & 46 & 11 & 15 & 0.77 & 0.57 & 0.81 \\
org:employees\_or\_members & 46 & 11 & 15 & 0.77 & 0.57 & 0.81 \\
per:neighbor & 40 & 14 & 12 & 0.67 & 0.73 & 0.65 \\
per:place\_of\_work & 37 & 9 & 25 & 0.62 & 0.47 & 1.35 \\
per:pet & 30 & 10 & 8 & 0.50 & 0.52 & 0.43 \\
per:acquaintance & 26 & 12 & 34 & 0.44 & 0.62 & 1.83 \\
per:origin & 21 & 4 & 1 & 0.35 & 0.21 & 0.05 \\
per:dates & 20 & 14 & 6 & 0.34 & 0.73 & 0.33 \\
per:schools\_attended & 5 & 2 & 1 & 0.08 & 0.10 & 0.05 \\
org:students & 5 & 2 & 1 & 0.08 & 0.10 & 0.05 \\
per:major & 2 & 1 & 3 & 0.03 & 0.05 & 0.16 \\
per:date\_of\_birth & 1 & 2 & 3 & 0.02 & 0.10 & 0.16 \\
gpe:birth\_in\_place & 0 & 1 & 0 & 0 & 0.05 & 0 \\
per:place\_of\_birth & 0 & 1 & 0 & 0 & 0.05 & 0 \\
\bottomrule
\end{tabular}
\caption{Statistics of relation labels in DialogRE dataset.}
\label{tab:relation-labels}
\end{table*}

\end{document}